\documentclass[runningheads]{llncs}
\usepackage[T1]{fontenc}
\usepackage{booktabs}
\usepackage[misc]{ifsym}

\usepackage{mwe}
\usepackage{graphicx}
\usepackage{dirtytalk}
\usepackage{soul}
\usepackage[hidelinks]{hyperref}
\usepackage[utf8]{inputenc}
\usepackage{amsmath}
\usepackage{amssymb}
\usepackage{bm}
\usepackage{booktabs}
\usepackage{xcolor}
\usepackage{yhmath}
\usepackage{bm}
\usepackage{bbm}

\graphicspath{ {figures/} }
 \usepackage{caption}
 \usepackage{subcaption}

\usepackage{newfloat}
\usepackage{listings}

\usepackage{mymacros}
\usepackage{tikz}
\usetikzlibrary{positioning,decorations.pathreplacing,shapes}

\newcommand{\argmax}{\arg\!\max}

\makeatletter
\newcommand\footnoteref[1]{\protected@xdef\@thefnmark{\ref{#1}}\@footnotemark}
\makeatother

\def\I{\mathrm{I}}

\usepackage{thmtools}
\usepackage{thm-restate}

\begin{document}
\title{\texorpdfstring{Understanding Domain-Size Generalization in\\ Markov Logic Networks}{Understanding Domain-Size Generalization in Markov Logic Networks}}

%
\author{ Florian Chen\inst{1}, Felix Weitkämper\inst{2} \and Sagar Malhotra\inst{1}}

\authorrunning{F. Chen  et al.}
%
\institute{TU Wien, Austria \and
 Ludwig-Maximilians-Universität München, Germany}
\maketitle              
\begin{abstract}
We study the generalization behavior of Markov Logic Networks (MLNs) across relational structures of different sizes. Multiple works have noticed that MLNs learned on a given domain generalize poorly across domains of different sizes. This behavior emerges from a lack of internal consistency within an MLN when used across different domain sizes. In this paper, we quantify this inconsistency and bound it in terms of the variance of the MLN parameters. The parameter variance also bounds the KL divergence between an MLN's marginal distributions taken from different domain sizes. We use these bounds to show that maximizing the data log-likelihood while simultaneously minimizing the parameter variance corresponds to two natural notions of generalization across domain sizes. Our theoretical results apply to Exponential Random Graphs and other Markov network based relational models. Finally, we observe that solutions that decrease MLN parameter variance, like regularization and Domain-Size Aware MLNs, increase the internal consistency of the MLNs. We empirically verify our results on four different datasets, with different methods to control parameter variance, showing that controlling parameter variance leads to better generalization. 
\end{abstract}


\section{Introduction}
Given the magnitude and ever-increasing nature of relational data, like social networks and epidemiology data, only a subsample of the data is ever observed. Statistical Relational Learning (SRL) \cite{SRL_LISA,SRL_LUC} methods integrate logic and probability to learn and infer over such data. However, \emph{are parameters estimated from subsampled data a good fit for the model of the larger relational structure?} Shalizi et al. \cite{Projectivity_Rinaldo} showed that, for most non-trivial probabilistic models on relations structures, it is \emph{probabilistically inconsistent} to apply the same model both to the whole relational structure and to its substructures. Jaeger et al. \cite{Projectivity_SRL} extend this analysis to a vast array of SRL models. These results show that, unlike independent and identically distributed (iid) data, relational data does not admit consistency of parameter estimation. That is, it is not true that the maximum likelihood (ML) parameter estimate converges to the true model parameters as the size of the observed data grows. In fact, these results show that the notion of a single true parameter, for relational structures of all sizes, is ill-defined for SRL models.

Lack of probabilistic consistency means that using an SRL model learned on a fixed domain for inference on a domain of different size may lead to poor results. The poor generalization behavior of SRL models across domain sizes is indeed observed in multiple empirical studies \cite{A_MLN,DA_MLN,Population_Extrapolation,ScaledStructureLearning}. Such issues can be ameliorated by using \emph{projective} models --- probabilistic models where the same parameters can be used for both the whole relational structure and its subsample. Formally, projective models capture probability distributions on relational structures (resp. graphs) of size $n$, where the marginal distribution over substructures (resp. subgraphs) of size $m < n$  does not depend on $n$. However, Shalizi et al. \cite{Projectivity_Rinaldo} also show that no projective model can express probability distributions with complex sufficient statistics, like $k$-cliques for any $k$ larger than two. These results also exclude the possibility of constructing any SRL model with practically desirable First-Order Logic (FOL) features such as transitivity.  Given these results, it is unclear what quantitative statements can be made about the generalization behavior of SRL models across domain sizes.

In this paper, we rigorously analyze domain-size generalization for a specific class of SRL models, namely Markov Logic Networks (MLNs). An MLN is a Markov Random Field with features defined in terms of weighted FOL formulas. We first formalize the notion of domain-size generalization of an MLN. We then provide an intuitive argument for what leads to non-projectivity, in terms of the weights that lead to dependence between the smaller and the larger domain. Theorem \ref{th: inequality_bounds} provides bounds on the difference between the probability distribution induced by an MLN on a subsampled domain and the probability distribution induced by the same MLN on a larger (unseen) domain.  We use this analysis to bound the KL divergence between the two distributions in terms of the parameter variance\footnote{We use the term \say{variance} in a colloquial sense here, as we actually bound the KL divergence in terms of the maximum and the minimum of the weight functions induced by an MLN} of the MLN. Finally, we show that maximizing the log-likelihood of an MLN on the subsampled domain, while minimizing the parameter variance, corresponds to (i) increasing the log-likelihood for generalization to the larger domain, and to (ii) reducing the KL divergence between the distributions induced by the MLN on the subsampled domain and the larger domain. Finally, we observe that methods like regularization and Domain-Size Aware MLNs \cite{DA_MLN} minimize the parameter variance, and hence lead to better generalization. We empirically verify these claims on four different datasets, with three different methods for controlling parameter variance.
Although the focus of this paper is on MLNs, our results can be generalized to Exponential Random Graph Models (ERGMs) and to any SRL model where template based parameter sharing is used \cite{PSL,Taskar2002DiscriminativePM}.

\section{Related work}


Lack of probabilistic consistency in probabilistic models on relational structures was first investigated by Shalizi et al. \cite{Projectivity_Rinaldo}. Jaeger et al. \cite{Projectivity_first} showed that such issues persist in most practically used SRL models. A large array of works have tried to devise new projective methods \cite{Projectivity_SRL,Projectivity_revisited} or identify and characterize projective fragments of existing SRL models \cite{Proj_MLN,Felix_Weit}, which circumvent these issues. However, most of the proposed new models are currently only of theoretical interest, as no clear way of learning or reasoning with them has been developed. On the other hand, most of the projective fragments of SRL models are rather restrictive.

These theoretical shortcomings are also reflected in the poor generalization behavior of SRL models in practice \cite{A_MLN,DA_MLN,Population_Extrapolation,ScaledStructureLearning}. Many works provide heuristic solutions \cite{A_MLN,DA_MLN,Population_Extrapolation} for better generalization across domain sizes. A particularly relevant family of formalisms adapts the parameter values with the size of the domain \cite{DA_MLN,DA-RLR}. In formalisms based on directed graphical models, it has been shown that parameter scaling leads to asymptotically projective models \cite{TypeIII}, 
but none of the heuristics developed for MLNs are formally motivated in this way \cite{DA-RLR}.
For MLNs, \cite{Relational_Marginal_Polytope} provides a sound approach for estimating the parameters for a larger unseen domain from a smaller subsample of fixed size. However, the practical applicability of this result is unclear. Furthermore, results provided in \cite{Relational_Marginal_Polytope} rely on learning on a larger domain. Hence, the computational complexity of learning can be prohibitively large in real-world settings.

In comparison to the aforementioned works, we analyze the generalization behavior of an MLN in the most natural setting, i.e., the MLN parameters are learned from a subsample of smaller size, and we analyze the behavior of such a distribution on the larger domain. Our analysis theoretically justifies many of the existing heuristic methods \cite{DA_MLN}. Our results are also relevant to works in ERGMs \cite{ERGM_PROJECTIVE}, investigating the relationship between the sample (resp. the substructure for us) and the population (resp. the larger relational structure for us).


\section{Background}
\subsection{Basic Definitions} The set of integers $\{1,...,n\}$ is denoted by $[n]$. We use $[m:n]$ to denote the set of integers $\{m,...,n\}$. Wherever the larger set of integers $[n]$ is clear from context, we will use $[\bar{m}]$ to denote the set $[m+1:n]$. For any $d \geq 1$, $\langle n \rangle^{d}$  represents $d$-tuples in $[n]^{d}$, with $d$ distinct elements, appearing in natural order. Hence, $\langle n \rangle^{d}$ forms a standardized representation of the set of all  $d$-element subsets in $[n]$. 

\subsection{First-Order Logic} We assume a function-free First-Order Logic (FOL) language
$\mathcal{L}$ defined by a finite set of variables $\mathcal{V}$, a finite set of symbols $\mathcal{R}$,  and a finite set of domain constants\footnote{Note that, w.l.o.g., we can assume the domain to be $[n]$ as we can always rename any finite domain of size $n$ with $[n]$.} $[n]$. For $a_1,...,a_k \in \mathcal{V}  \cup [n]$ and $R\in \mathcal{R}$ we call $R(a_1,...a_k)$ an \emph{atom}.  If $a_1,...,a_k \in [n]$, then the atom $R(a_1,...a_k)$  is called a \emph{ground atom}. A \emph{literal} is an atom or the negation of an atom. We assume \emph{Herbrand Semantics} \cite{Herbrand_Logic}. Hence, a \emph{world} or an \emph{interpretation} is simply a mapping of each ground atom to a boolean.  The set of interpretations over a domain of size $n$ is denoted by $\Omega^{(n)}$. For a subset $\mathrm{I}\subset [n]$ we use $\omega \downarrow \mathrm{I}$ to denote the partial interpretation induced by $\mathrm{I}$. Thus, $\omega \downarrow \mathrm{I}$  is an interpretation over the ground atoms containing only the domain constants in $\mathrm{I}$. For any $\mathbf{c} \in \langle n \rangle^{d}$ we use $\omega \downarrow \mathbf{c}$ to denote the partial interpretation induced by the domain elements in the tuple $\mathbf{c}$. 




\begin{example}
    \label{ex: projection}
    Consider a formal language comprising only two binary relation symbols, denoted as $G$ and $B$. We can visualize an interpretation $\omega$ as a multi-relational directed graph. In this graph, a directed edge of color green (for $G$) or blue (for $B$) connects two nodes $x$ and $y$ if and only if $G(x,y)$ or $B(x,y)$ respectively holds true in $\omega$. For an illustrative interpretation $\omega$ on the set $\Delta = [4]$, the graphical representation is as follows:
\noindent
\begin{center}
\begin{tikzpicture}[node distance={13mm}, thick, main/.style = {draw, circle}] 
    \node[main] (1) {1}; 
    \node[main] (2) [right of=1] {2}; 
    \node[main] (3) [below of=1] {3}; 
    \node[main] (4) [below of=2] {4};  
    \draw[green!70!black!70, ->,line width=2pt] (2) -- (1); 
    \draw[blue,->,line width=2pt] (2) to [out=270,in=90,looseness=1] (4);
    \draw[green!70!black!70,->,line width=2pt] (3) to [out=225,in=315,looseness=5] (3);
    \draw[blue,->,line width=2pt] (3) -- (4); 
        \end{tikzpicture}
    \end{center}
\noindent
Then, the two subsets $\omega' = \omega \downarrow [2]$ and $\omega'' = \omega \downarrow [\bar{2}]$ can graphically be represented as
\noindent
\begin{center}
    \begin{tikzpicture}[node distance={13mm}, thick, main/.style = {draw, circle}] 
    \node[main] (1) {1}; 
    \node[main] (2) [right of=1] {2}; 
    \node[main] (3) [right=5cm of 2] {3}; 
    \node[main] (4) [right of=3] {4};  
    \draw[green!70!black!70,->,line width=2pt] (2) -- (1);
    \draw[green!70!black!70,->,line width=2pt] (3) to [out=230,in=310,looseness=5] (3); 
    \draw[blue,->,line width=2pt] (3) -- (4); 
            \end{tikzpicture}
\end{center}
\end{example}
Note that if $\mathbf{c} = \langle 1,2 \rangle$, then $\omega \downarrow \mathbf{c} = \omega \downarrow [2]$.

\subsubsection{Families of Probability Distributions} We will deal with probability distributions on a set of interpretations. A family of probability distributions $\{P^{(n)}: n \in \mathbb{N}  \}$ specifies, for each finite domain of size $n$, a distribution $P^{(n)}$ on the possible $n$-world set $\Omega^{(n)}$ \cite{Projectivity_SRL}. We will work with  \emph{exchangeable} probability distributions  \cite{Projectivity_SRL}. These are distributions where $P^{(n)}(\omega) = P^{(n)}(\omega') $ if $\omega$ and $\omega'$ are isomorphic. A distribution $P^{(n)}(\omega)$ over $n$-worlds induces a marginal probability distribution over  $m$-worlds $\omega' \in \Omega^{(m)}$, where $m \leq n$, as follows:
\begin{equation*}
    P^{(n)}\downarrow [m](\omega') = \sum_{\omega \in \Omega^{(n)}:\omega \downarrow [m] =\omega'} P^{(n)}(\omega)
\end{equation*}
Note that due to exchangeability $P^{(n)} \downarrow \mathrm{I}$ is the same for all subsets $\I$ of size $m$. Hence, we can always assume any induced $m$-world  to be $\omega \downarrow [m]$. We can now  define projectivity as follows:
\begin{definition}[Projectivity \cite{Projectivity_SRL}]
    An exchangeable family of probability distributions is called projective if for all $m < n$: $$P^{(n)} \downarrow [m] = P^{(m)}$$
\end{definition}
\noindent

\section{Learning in Markov Logic}
\label{sec:learning}
A Markov Logic Network (MLN) $\Phi$ is defined by a set of weighted
formulas $\{(\phi_i,a_i)\}_i$, where $\phi_i$ are function-free, quantifier-free, FOL formulas with weights $a_i \in \mathbb{R}$. An MLN $\Phi$ induces a
probability distribution over the set of interpretations $\Omega^{(n)}$: 
\begin{equation}
\label{eq: MLN}
    P^{(n)}_{\Phi}(\omega) = \frac{1}{Z(n)}\exp\Bigl(\sum_{(\phi_i,a_i)\in \Phi}a_i N(\phi_i,\omega)\Bigr)
\end{equation}
where $N(\phi_i,\omega)$ is the number of true groundings of
$\phi_i$ in $\omega$. The normalization constant $Z(n)$ is called the
\emph{partition function} that ensures that $P^{(n)}_{\Phi}$ is a
probability distribution. In the following, we provide an example of an MLN which models the spread of COVID-19 due to contact among different individuals and the impact of vaccines. 

\begin{example} 
\label{Ex: MLN_Example}
Let us have a relational language with the unary predicates $\mathtt{Covid}$ and $\mathtt{Vaccine}$, and a binary predicate $\mathtt{Contact}$. An MLN can be defined as follows:
\begin{align*}
    a_1 &\quad \mathtt{Vaccine}(x) \rightarrow  \mathtt{ \neg Covid}(x)\\
    a_2 &\quad \mathtt{Covid}(x) \land \mathtt{Contact}(x,y) \rightarrow  \mathtt{Covid}(y)
\end{align*}
\end{example}

Like in most SRL models, learning in MLNs is guided by the maximum likelihood (ML) principle. Formally, given an observed relational structure $\omega \in \Omega^{(n)}$, and an MLN $\Phi$, the ML estimate for the weights is given follows:
\begin{equation}
    \label{Eq: ML_estimate}
    \mathbf{\hat{a}} = \argmax_{\mathbf{a}} P^{(n)}_{\Phi}(\omega) 
\end{equation}
where $P^{(n)}_{\Phi}(\omega)$ is the probability distribution due to an MLN on the set of interpretations $\Omega^{(n)}$ as defined in equation \eqref{eq: MLN}. However, in most cases, the observed relational structure $\omega$ is a substructure of some larger structure on a larger unobserved domain. For instance, the number of people tested during a pandemic, and the number of contacts reported (say, using a contact-tracing mobile application) are only a subset of the true infection-contact network, which is spread over the entire local or even global population. Hence, our goal is to estimate the parameters for the MLN distribution $P^{(n+m)}_{\Phi}$ for some (potentially very large) $m$ using only the substructure $\omega$ of size $n$. Formally, we want the following ML estimate: 
\begin{equation}
    \label{Eq: ML_estimate_m_n}
    \mathbf{\hat{a}} = \argmax_{\mathbf{a}} P^{(n+m)}_{\Phi} \downarrow [n](\omega) 
\end{equation}

However, given that most MLNs are not projective \cite{Projectivity_SRL,Projectivity_Rinaldo}, the ML estimate in equation \eqref{Eq: ML_estimate_m_n} is not the same as the ML estimate in \eqref{Eq: ML_estimate}. As $m$ may be very large, it can be computationally prohibitive to make the ML estimate for the distribution $P^{(n+m)}_{\Phi}$. Furthermore, in many cases it may be hard to know or guess the value of $m$. Hence, our goal would be to analyze the relation between the distributions $P^{(n)}_{\Phi}(\omega)$ and $P^{(n+m)}_{\Phi} \downarrow [n](\omega)$, and use that analysis to subsequently characterize conditions that lead to better ML parameter estimates for $P^{(n+m)}_{\Phi}$, or in other words generalize better to larger domains.



\begin{remark}
Although projective MLNs can easily be obtained, their expressivity is significantly limited. One projective fragment of MLNs is the $\sigma$-determinate MLNs \cite{Projectivity_SRL,MLN_Inf_Dom}. A Markov Logic Network $\Phi := \{\phi_i,a_i\}_i$ is $\sigma$-determinate if its formulas $\phi_i$ satisfy that any two atoms appearing in $\phi_i$ contain the same variables.
\end{remark}
\begin{example} 
\label{Ex: Sigma_Determinate}
Following is an example of a $\sigma$-determinate MLN:
\begin{align*}
    a_1 &\quad \mathtt{Covid}(x) \\
    a_2 &\quad \mathtt{Contact}(x,y) \land \mathtt{Contact}(y,x)
\end{align*}
\end{example}
\noindent
 Example \ref{Ex: Sigma_Determinate} shows  a $\sigma$-determinate MLN. Even simple MLNs, such as the one presented in Example \ref{Ex: MLN_Example}, can not be represented as a $\sigma$-determinate MLN. Hence, our goal in this paper can also be framed as to obtain arbitrarily expressive MLNs that are close to being projective. 
\section{Markov Logic Across Domain Sizes}
  
In this section, we analyze how the weights induced by an MLN distribute over different parts of the domain. We present the necessary machinery for our main results in Section \ref{sec: main_results} and create an intuition for what leads to projectivity, and how any MLN can be made \emph{closer} to being projective.

We assume, w.l.o.g., that each $k$-ary formula in an MLN can be grounded only to $k$ distinct domain constants. This does not restrict the expressivity of an MLN, as an MLN with a formula $\psi(x,y)$ with weight $a$ can be equivalently expressed by replacing $\psi(x,y)$ with two formulas: $\psi(x,x)$ and $\psi(x,y) \land (x \neq y)$ with the same weight $a$. This principle can be generalized to formulas with arbitrary arity. We will use $\Phi_k$ to represent the subset of weighted formulas in an MLN $\Phi$ with arity $k$. We now define weight functions for a given MLN $\Phi$. 


    

\begin{definition}[weight function] Given an MLN $\Phi$, we define the weight of an interpretation $\omega$ as follows:
\begin{equation}
\label{eq: weight}
    w(\omega) = \exp\Bigl(\sum_{(\phi_i,a_i)\in \Phi}a_i N(\phi_i,\omega)\Bigr)
\end{equation}
\end{definition}

We will also need to decompose the weight contribution of different $k$-tuples to the weight $w(\omega)$. To that end, we define the $k$-weight functions as follows:
\begin{definition}[$k$-weight function] Given an MLN $\Phi$, we define the $k$-weight of an interpretation $\omega$ as follows:
\begin{equation}
    w_k(\omega) = \exp\Bigl(\sum_{(\phi_i,a_i)\in \Phi_k}a_i N(\phi_i,\omega)\Bigr)
\end{equation}
where $\Phi_k$ is the subset of  weighted formulas in $\Phi$ with arity $k$.
\end{definition}

 In the following two Lemmas, we further decompose the contribution of each $k$-substructure towards the weight $w(\omega)$. 

\begin{lemma}
\label{lem: model_weight}
Given an MLN with weight function $w$ and $k$-weight functions $w_k$, then:
\begin{equation}
\label{eq: k_weight}
   w(\omega) = \prod_{k \in [d]} \prod_{\mathbf{c} \in \langle n \rangle^{k}}w_{k}(\omega \downarrow \mathbf{c})
\end{equation}
where $d$ is the largest arity of the formulas in the MLN.
\end{lemma}
\begin{proof}
Let $\phi \in \Phi_k$ be an arbitrary weighted formula with $k$ variables. The  weight contribution of $\phi$ to $\sum_{(\phi_i,a_i)\in \Phi}a_i N(\phi_i,\omega)$ is given by the weighted number of true groundings of $\phi$ in $\omega$. Since $\phi$ is always grounded to distinct domain constants and it has arity $k$, its weight contribution is the sum of its weight contribution to each of the $\omega \downarrow \mathbf{c}$ for $\mathbf{c} \in \langle n \rangle^{k}$. Repeating the same argument for all arities and all formulas in the MLN, we have that:

\begin{equation*}
    \sum_{(\phi_i,a_i)\in \Phi}a_i N(\phi_i,\omega) =  \sum_{k \in [d]}\sum_{\mathbf{c} \in \langle n \rangle^{k}}\sum_{(\phi_i,a_i)\in \Phi_k}a_i N(\phi_i,\omega \downarrow \mathbf{c})
\end{equation*}
Hence, 
\begin{align*}
    \exp \left(\sum_{(\phi_i,a_i)\in \Phi}a_i N(\phi_i,\omega)\right) &=  \exp \left( \sum_{k \in [d]}\sum_{\mathbf{c} \in \langle n \rangle^{k}}\sum_{(\phi_i,a_i)\in \Phi_k}a_i N(\phi_i,\omega \downarrow \mathbf{c}) \right) \\
 &= \prod_{k \in [d]} \prod_{\mathbf{c} \in \langle n \rangle^{k}}w_{k}(\omega \downarrow \mathbf{c})
\end{align*}
\qed
\end{proof}
Similar weight functions can be constructed for other Markov network based SRL models \cite{PSL,Taskar2002DiscriminativePM} where template based parameter sharing is used. 
\begin{lemma}
\label{lem: model_split}
If $\omega$ is an interpretation on a domain $[n+m]$, then $w(\omega)$ can be factorized as follows:   
\begin{equation}
\label{eq: model_split}
w(\omega) = w(\omega \downarrow [n]) \times w(\omega \downarrow [\bar{n}]) \times \prod_{k \in [d]} \prod_{\mathbf{c} \in \langle n + m \rangle^{k} \setminus  \langle n \rangle^{k} \cup \langle \bar{n} \rangle^{k}  } \!\!\!\! w_{k}( \omega \downarrow \mathbf{c})
\end{equation}
\end{lemma}
\begin{proof} 
\begin{align*}
&w(\omega) = \prod_{k \in [d]} \prod_{\mathbf{c} \in \langle n+m \rangle^{k}} w_{k}( \omega \downarrow \mathbf{c})\\
&= \prod_{k \in [d]} \prod_{\mathbf{c} \in \langle n\rangle^{k}} w_{k}( \omega \downarrow \mathbf{c}) \prod_{k \in [d]} \prod_{\mathbf{c} \in \langle m \rangle^{k}} w_{k}( \omega \downarrow \mathbf{c}) \prod_{k \in [d]} \prod_{\mathbf{c} \in \langle n + m \rangle^{k} \setminus \langle n \rangle^{k} \cup \langle \bar{n} \rangle^{k}  } \!\!\!\!\!\!\!\! w_{k}( \omega \downarrow \mathbf{c})\\
&= w(\omega \downarrow [n]) \times w(\omega \downarrow [\bar{n}]) \times \prod_{k \in [d]} \prod_{\mathbf{c} \in \langle n + m \rangle^{k} \setminus \langle n \rangle^{k} \cup \langle \bar{n} \rangle^{k}  } \!\!\!\! w_{k}( \omega \downarrow \mathbf{c})
\end{align*}
\qed
\end{proof}

\noindent
A key part of our analysis would be understanding the weight contribution to the probability distribution $P^{(n+m)}_{\Phi}\downarrow [n] (\omega')$ due to the following term in Lemma \ref{lem: model_split}:
\begin{equation}
\label{eq: cross_term}
    \prod_{\mathbf{c} \in \langle n + m \rangle^{k} \setminus  \langle n \rangle^{k} \cup \langle \bar{n} \rangle^{k}  } \!\!\!\! w_{k}( \omega \downarrow \mathbf{c})
\end{equation}
Expression \eqref{eq: cross_term} captures weight contribution from $k$-tuples which are strictly not part of the domain $\langle n  \rangle^{k}$, and neither of the domain $\langle \bar{n} \rangle ^{k}$. Intuitively, our goal is to control the weight contributions due to the relations that create dependence between the observed relational structure on the domain $[n]$, and the unobserved relational structure on the domain $[\bar{n}]$.

\section{Domain-Size Generalization}
\label{sec: main_results}

In this section, we present the main results of our paper. Let $w_k^{max}$ and $w_k^{min}$ denote the maximum and the minimum of the weight function $w_k$. 

\begin{proposition} 
\label{prop: factorising_interpetation}

Given an interpretation $\omega$ on the domain $[n+m]$, then

\begin{align}
w(\omega) &\leq w(\omega \downarrow [n]) \times w(\omega \downarrow [\bar{n}]) \times \prod_{k \in [d]} (w^{max}_{k})^{ \binom{n+m}{k} - \binom{n}{k} - \binom{m}{k}}\\
w(\omega) &\geq w(\omega \downarrow [n]) \times w(\omega \downarrow [\bar{n}]) \times \prod_{k \in [d]} (w^{min}_{k})^{ \binom{n+m}{k} - \binom{n}{k} - \binom{m}{k}}
\end{align}
\end{proposition}

\begin{proof}
    The statement follows from equation \eqref{eq: model_split} in Lemma \ref{lem: model_split}. The upper bound is obtained by replacing the multiplicative weight contribution of each tuple in $\langle n + m \rangle^{k} \setminus \langle n \rangle^{k} \cup \langle \bar{n} \rangle^{k}$ with $w_k^{max}$, for all $k \in [d]$. And the lower bound is obtained by replacing the weight contribution of all such tuples with $w_k^{min}$.
    \qed
\end{proof}

For ease of notation, we define the following new parameters: 
\begin{align}
    M_{max} &= \prod_{k \in [d]}  (w^{max}_{k})^{ \binom{n+m}{k} - \binom{n}{k} - \binom{m}{k}} \label{eq: m_max}\\
    M_{min} &= \prod_{k \in [d]} (w^{min}_{k})^{ \binom{n+m}{k} - \binom{n}{k} - \binom{m}{k}} \label{eq: m_min}
\end{align}
     
\begin{proposition}
\label{prop: non-projective mln meets bounds}
    There exists an MLN for which the bounds in Proposition \ref{prop: factorising_interpetation} are met for some interpretation $\omega$.
\end{proposition}

\begin{proof} 
Assume an MLN with only the formula $R(x,y) \land R(y,z) \land R(x,z)$, with weight $a > 0$. It can be checked that the upper bound is met for an $\omega \in \Omega^{(n+m)}$ where all the domain constants are related w.r.t. the relation $R$. And the lower bound is met by the $\omega' \in \Omega^{(n+m)}$, such that no relation between any of the domain constants exist.
\qed
\end{proof}
\noindent
Proposition \ref{prop: non-projective mln meets bounds} shows that bounds in Proposition \ref{prop: factorising_interpetation} can not be improved. 

\begin{proposition}
\label{prop: Z factorisation}
Given a Markov Logic Network, we have that 
 \begin{equation}
    M_{min} C_{n,m}  Z(n)Z(m) \leq Z(n+m) \leq Z(n)Z(m)C_{n,m}  M_{max} 
 \end{equation}
where $C_{n,m}$ is the number of ways in which an interpretation on $[n]$ and an interpretation on $[\bar{n}]$ can be extended to an interpretation on $[n+m]$.
\end{proposition}
\begin{proof}
\begin{align*}
Z(n+m) &=  \sum_{\omega } w(\omega) \\
&\leq \sum_{\omega} w(\omega \downarrow [n]) \times w(\omega \downarrow [\bar{n}]) \times \prod_{d \in [k]} (w^{max}_{d})^{ \binom{n+m}{d} - \binom{n}{d} - \binom{m}{d}} \\
&= \sum_{\omega} w(\omega \downarrow [n]) \times w(\omega \downarrow [\bar{n}]) \times M_{max} \\
& = M_{max} \sum_{\substack{\omega' \in \Omega^{(n)} \\ \omega'' \in \Omega^{(m)}}} C_{n,m} \times  w(\omega') \times w(\omega'')\\
&= M_{max} C_{n,m}Z(n)Z(m)
\end{align*}
\qed
\end{proof}
\noindent
As the proof of the lower bound follows analogous to the proof of the upper bound, we defer it to the appendix. 

 We now present the main result of the paper:
\begin{theorem} Given a Markov Logic Network $\Phi$, then the following inequality holds for all $\omega \in \Omega^{(n)}$:
\label{th: inequality_bounds}
\begin{equation}
\label{eq: inequality_bounds}
    \frac{M_{min}}{M_{max}}  P_{\Phi}^{(n)}(\omega) \leq P_{\Phi}^{(n+m)}\downarrow [n] (\omega) \leq \frac{M_{max}}{M_{min}}  P_{\Phi}^{(n)}(\omega)
\end{equation}
\end{theorem}

\begin{proof}
\begin{align*}
      P_{\Phi}^{(n+m)}\downarrow [n] (\omega') &= \sum_{\substack{  \omega \in \Omega^{(n+m)} \\ \omega \downarrow [n] = \omega'}} \frac{w(\omega)}{Z(n+m)}
\end{align*}
Using Proposition \ref{prop: Z factorisation}, we have:
\begin{align*}
      P_{\Phi}^{(n+m)}\downarrow [n] (\omega') &\leq \frac{1}{Z(n)Z(m)M_{min}C_{n,m}} \sum_{\substack{  \omega \in \Omega^{(n+m)} \\ \omega \downarrow [n] = \omega'}} w(\omega)
\end{align*}
Using Proposition \ref{prop: factorising_interpetation}, we have:
\begin{align*}
      P_{\Phi}^{(n+m)}\downarrow [n] (\omega') &\leq   \frac{1}{Z(m)M_{min} C_{n,m}} \sum_{\substack{  \omega \in \Omega^{(n+m)} \\ \omega \downarrow [n] = \omega'}} \frac{w(\omega') w(\omega \downarrow [\bar{n}]) M_{max}}{Z(n)}\\
   \end{align*}
Hence, we have that
   \begin{align*}
      P_{\Phi}^{(n+m)}\downarrow [n] (\omega') &\leq \frac{1}{Z(m)M_{min}C_{n,m}} w(\omega') \frac{M_{max}}{Z(n)} \sum_{\omega'' \in \Omega^{(m)}} \sum_{\substack{  \omega \in \Omega^{(n+m)} \\ \omega \downarrow [n] = \omega' \\ \omega \downarrow [\bar{n}] = \omega''}} w(\omega'')\\
       &=\frac{1}{Z(m)M_{min}C_{n,m}} w(\omega') \frac{M_{max}}{Z(n)} \sum_{\omega'' \in \Omega^{(m)}} C_{n,m} w(\omega'') \\
        &=\frac{1}{Z(m)M_{min}C_{n,m}} w(\omega') \frac{M_{max}}{Z(n)} C_{n,m} Z(m)\\
      &=  \frac{M_{max}}{M_{min}}  P^{(n)}_{\Phi}(\omega')
   \end{align*}
  \qed
\end{proof}
The proof of the lower bound follows analogous to the proof of the upper bound, we defer it to the appendix. Let us now denote $\frac{M_{max}}{M_{min}}$ with the symbol $\Delta$.



\begin{corollary}
\label{cor: log-likelihood + sigma}
 $$  -\log P_{\Phi}^{(n+m)}\downarrow [n](\omega) \leq -\log P_{\Phi}^{(n)}(\omega)  + \log\Delta  $$
\end{corollary}
Corollary \ref{cor: log-likelihood + sigma} is a simple consequence of Theorem \ref{th: inequality_bounds} and its proof is therefore deferred to the appendix.

 Corollary \ref{cor: log-likelihood + sigma} shows that minimizing the negative log-likelihood of the observed subsample $\log P_{\Phi}^{(n)}(\omega)$, while simultaneously reducing $\log \Delta$, leads to the upper bound on the negative marginal log-likelihood being reduced. Hence, bringing the parameter estimate closer to the ML estimate as required by equation \eqref{Eq: ML_estimate_m_n}. Note that the ML estimate in equation \eqref{Eq: ML_estimate_m_n} takes into account that the observed structure is a subsample of a larger relational structure, and optimizes the weights to get the best estimate for the larger domain size.

As $\Delta$ is the quotient of $M_{max}$ and $M_{min}$, as defined in equations \eqref{eq: m_max} and \eqref{eq: m_min}, reducing $\log \Delta$ corresponds to reducing the difference between the largest and the smallest values taken by $\log w_k$. This can be easily achieved by a simple regularization objective on the weights $a_i$ of the MLN.

\begin{theorem}
\label{th: kl-divergence}
    $$  KL(P^{(n+m)}_{\Phi}\downarrow [n] || P^{(n)}_{\Phi}) \leq \log \Delta $$
\end{theorem}
\begin{proof}
\begin{align*}
KL(P^{(n+m)}_{\Phi}\downarrow [n] || P^{(n)}_{\Phi}) &= \underset{\omega \in \Omega^{(n)}}{\sum} P^{(n+m)}_{\Phi}\downarrow [n](\omega) \times \log\left(\frac{P^{(n+m)}_{\Phi}\downarrow [n](\omega)}{P^{(n)}_{\Phi}(\omega)}\right)\\
&\leq \underset{\omega \in \Omega^{(n)}}{\sum} P^{(n+m)}_{\Phi}\downarrow [n](\omega) \times \log\left(\frac{\Delta \times  P^{(n)}_{\Phi}(\omega)}{P^{(n)}_{\Phi}(\omega)}\right)\\
&= \underset{\omega \in \Omega^{(n)}}{\sum} P^{(n+m)}_{\Phi}\downarrow [n](\omega) \times \log \Delta
\end{align*}
Note that $$\underset{\omega \in \Omega^{(n)}}{\sum} P^{(n+m)}_{\Phi}\downarrow [n](\omega) \times \log \Delta$$ is the expectation value of $\log \Delta$ under the distribution  $P^{(n+m)}_{\Phi}\downarrow [n](\omega)$. Since the expectation of a constant is the constant itself, we have that:
$$\underset{\omega \in \Omega^{(n)}}{\sum} P^{(n+m)}_{\Phi}\downarrow [n](\omega) \times \log \Delta = \log \Delta$$
\qed
\end{proof}

Theorem \ref{th: kl-divergence} gives an easy method of minimizing the upper-bound, on the otherwise intractable, KL-divergence between $P^{(n+m)}_{\Phi}\downarrow [n]$  and $ P^{(n)}_{\Phi}$. Hence, an MLN learning procedure can be pushed to have smaller $KL(P^{(n+m)}_{\Phi}\downarrow [n] || P^{(n)}_{\Phi})$, and in turn be incentivized towards representing a projective distribution, simply by minimizing the $\log \Delta$ term i.e. the difference between the minimum and the maximum of the $k$-weight functions.

\begin{corollary}
\label{cor: kl-divergence and likelihood}
    $$  -\log P^{(n)}_{\Phi}(\omega) +  KL(P^{(n+m)}_{\Phi}\downarrow [n] || P^{(n)}_{\Phi}) \leq -\log P^{(n)}_{\Phi}(\omega)  + \log \Delta    $$
\end{corollary}
This statement can easily be derived from Theorem \ref{th: kl-divergence}. We defer its proof to the appendix.

Corollary \ref{cor: kl-divergence and likelihood} characterizes another notion of generalization across varying domain sizes. By minimizing the negative log-likelihood and the difference between $M_{min}$ and $M_{max}$, we have that the upper-bound on the negative log-likelihood plus the KL divergence between the two distributions is minimized. This minimization can be seen as optimizing a dual objective. On the one hand, the likelihood of the observed substructure is maximized w.r.t. the distribution $P^{(n)}_{\Phi}$. While on the other hand,  $P^{(n)}_{\Phi}$ is moved closer to $P^{(n+m)}_{\Phi}\downarrow [n]$ in terms of KL-divergence. This minimization of KL divergence can be seen as incentivizing distributions which are \emph{closer} to being projective.

\section{Experiments}

In this section, we evaluate the effect of reducing parameter variance on generalization behavior\footnote{\label{note: regularization_code}Our code is available \href{https://github.com/faguodev/regularization_experiments}{Online}.}
Proposition \ref{prop: factorising_interpetation} bounds $w(\omega)$ w.r.t. the maxima and minima of $w_{k}$. However, for almost all worlds, this bound is loose. This is because, for most worlds, not all $k$-tuples chosen from across the domains will have the extreme weights. Note that in general, our goal is to minimize the impact of the term presented in equation \eqref{eq: cross_term}. Therefore, it is more effective to reduce the spread between all the weights, rather than merely scaling the upper and the lower bound. Also note that for most MLNs, for some $\omega \in \Omega^{k}$, we will have that $w_{k}(\omega)=1$, i.e., none of the formulas in the MLN will be realized on $\omega$. Thus, in most practical cases,  to reduce the spread of the weights $a_i$, one should reduce their spread around $0$.

Multiple approaches discussed in the literature, directly or indirectly, minimize the parameter variance  \cite{huynh2008discriminative,DA_MLN}. We empirically evaluate the effects of three such approaches: L1 regularization, L2 regularization, and Domain-Size Aware Markov Logic Networks (DA-MLNs) \cite{DA_MLN}. 
Both L1 and L2 regularization directly work to reduce the spread of the parameters: L1 regularization penalizes the sum of the absolute weight values and L2 regularization penalizes the sum of squared weights. In our setting, we only penalize formulas of arity $>1$, because unary formulas do not affect the connecting term discussed in equation \eqref{eq: cross_term}. 

A DA-MLN is an adaptation of a regular MLN that reduces the variance of the parameters by down-scaling formula weights depending on the domain size of the dataset it should generalize to. In this section, we will call such datasets \emph{target sets}. A DA-MLN is then given as follows:

\begin{equation}
    P^{(n)}_{\Phi}(\omega) = \frac{1}{Z(n)}\exp\Bigl(\sum_{(\phi_i,a_i)\in \Phi}\frac{a_i}{s_i} N(\phi_i,\omega)\Bigr)
\end{equation}
The scale-down factor $s_i$ is defined as follows:
\begin{align}
   s_i =  \underset{P \in \phi_i}{\max}\Bigl( \max\bigl(1,\underset{x \in Vars_{i}(P)^-}{\prod} \lvert\Delta_x\rvert\bigr)\Bigr)
\end{align}
where $\lvert \Delta_x \rvert$ is the domain size of $x$ in the target set and $Vars_{i}(P)^-$ is the set of logical variables appearing in $\phi_i$ but not in the atom $P$.

To precisely verify our theoretical results, we employ Lifted Inference \cite{kersting2012lifted,First_Order_Prob_Inf} and Lifted Generative Learning \cite{vanHaaren2016}. These methods allow us to compute and compare exact dataset likelihoods. In contrast, alternative methods optimize approximate objectives, such as pseudo-likelihood \cite{besag1975statistical}, which may interfere with the verification of the theoretical results. However, using lifted methods restricts the expressivity of the MLNs we can test.


\subsection{Datasets}

To provide a thorough analysis of the effects of different methods for generalizing across different domain sizes, we use four datasets commonly used in related literature: Friends \& Smokers (FS) \cite{singla2008}, IMDB\footnote{\label{note2}Dataset available on the \href{https://alchemy.cs.washington.edu/data/}{Alchemy website.}} \cite{mihalkova2007}, WebKB\footnoteref{note2} \cite{mihalkova2007} and Nations\footnoteref{note2} \cite{rummel1992}.\\
\\
\noindent \textbf{Friends \& Smokers (FS).} This synthetic dataset captures information about smoking habits, friendships, and cancer diagnoses of a set of people. The data is created by first randomly selecting 40\% of a population to be smokers. Then, 30\% of the smokers and 10\% of the non-smokers are chosen to suffer from cancer. Lastly, friendships are assigned based on smoking habits, with a $0.8$ probability for friendships between people with the same smoking habit, and a $0.1$ probability of friendships between people with different smoking habits. For our experiments, we generate a target set of size 500.\\
\\
\noindent \textbf{IMDB.} Taken from the International Movie Database this dataset contains information about movies and, their actors and directors. Also included are certain attributes like gender and work relations of actors and directors. The dataset has a total of 297 constants, of which 268 are of type person, 20 are of type movie, and 9 are of type genre. The dataset contains 3 binary and 3 unary predicates.\\
\\
\noindent \textbf{WebKB} This dataset captures information about web pages from four US universities. For each web page, the original dataset \cite{craven2001} includes a label (e.g. Course, Faculty) as well as textual information about the page contents. Similar to Mihalkova et al. \cite{mihalkova2007}, the version we use disregards the textual information and focuses on page classes and relations, for example between courses and teaching assistants. This version of the dataset comprises a total of 989 constants, of which 746 are of type person. The dataset contains 3 binary and 2 unary predicates.\\
\\
\noindent \textbf{Nations.} This dataset contains a set of features of nations and relations between them. Relations include treaties and (economic-)aid, features include governance types and technological advancements. In total, there are 14 nations, 111 features (given as unary predicates), and 56 relations (given as binary predicates).

\subsection{Methodology}
We compare the generalization behavior of standard generative weight learning to different methods that also reduce parameter variance: L1 regularization, L2 regularization, and DA-MLNs. The structures of the MLNs we use are adopted from Van Haaren et al. \cite{vanHaaren2016}, who introduced a Lifted Structure Learning (LSL) approach. LSL ensures that the learned structures are liftable and learnable in practice. The Nations dataset, with over 160 predicates, presents an infeasibly large search space of possible clauses for LSL. Hence, we use a hand-crafted MLN of 50 formulas. For weight learning, we employ Lifted Generative Learning \cite{vanHaaren2016}. This allows us to compare the exact target set likelihoods, which is the natural evaluation measure for generative learning \cite{darwiche2009modeling,koller2009probabilistic,murphy2012machine,vanHaaren2016} and is also best suited for validating our theoretical results.

To provide reliable results we generate 20 training set for weight learning and 5 target sets for each of the sizes we want to generalize to. For generating a training set, we uniformly sample a subset from a specific type $\tau$ of constant: We sample 20 \emph{persons} for FS, 50 \emph{persons} for IMDB and WebKB, and 5 \emph{nations} for the Nations dataset. Now, let $\mathrm{I}$ denote the set of the sampled constants. In the training set, we then include all the ground atoms $R(a_1,...a_k)$ where all the domain constants of type $\tau$ in  $\{a_1,...,a_k\}$ are included in $\mathrm{I}$. The process for generating target sets follows a similar approach. 

For standard generative weight learning and DA-MLNs, we then learn the weights on each training set and compute the log-likelihood of each target set. For L1 and L2 regularization, to find the best regularization parameter $\lambda$, we perform hyperparameter tuning on the values between $10^{-2}$ and $10^2$ on the smallest target sets\footnote{Hence, the results on the smallest target sets are slightly biased for L1 and L2. However, for larger target set sizes, no such bias exists.}. As our metric to compare the different approaches, we measure how much the target set log-likelihood improves in comparison to no regularization. This metric measures how well our ML estimate is w.r.t. equation \eqref{Eq: ML_estimate_m_n}, i.e., the ML estimate that takes into account the fact that the observed data came from a larger relational structure.

\subsection{Results}

\begin{figure}
     \centering
     \begin{subfigure}[b]{0.49\textwidth}
         \centering
         \includegraphics[width=\textwidth]{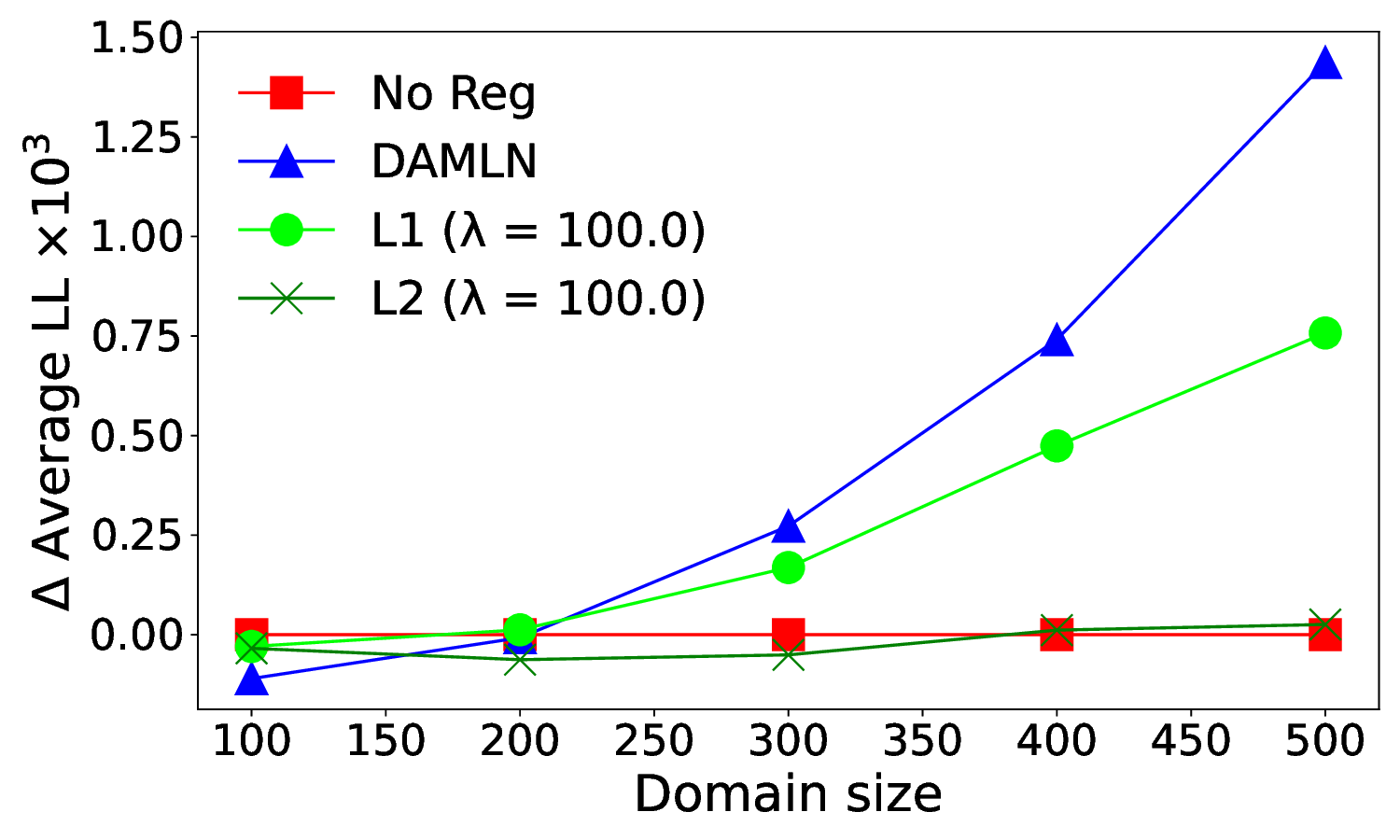}
         \caption{$\Delta$-log-likelihoods (FS)}
         \label{Log-likelihoods (FS)}
     \end{subfigure}
     \hfill
     \begin{subfigure}[b]{0.49\textwidth}
         \centering
         \includegraphics[width=\textwidth]{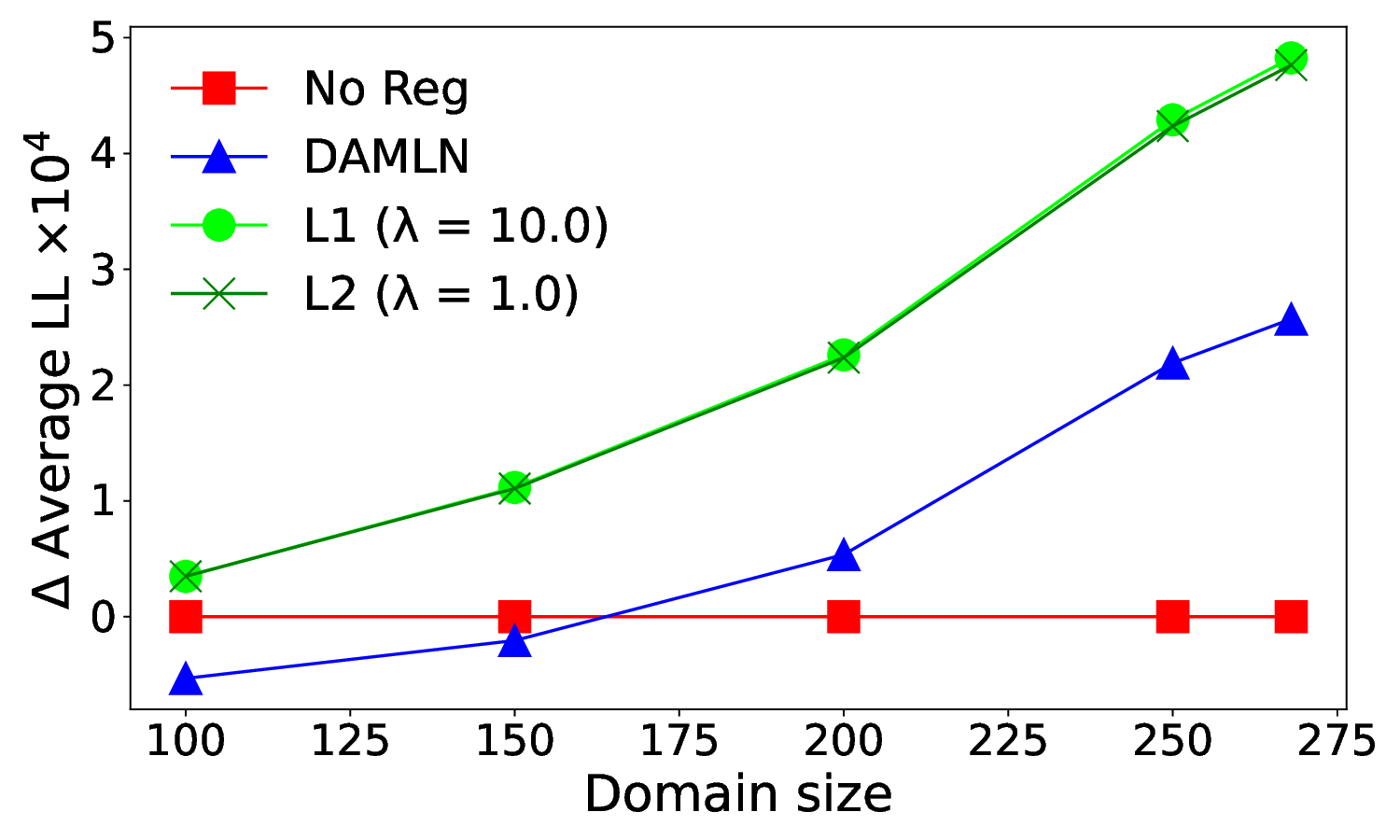}
         \caption{$\Delta$-log-likelihoods (IMDB)}
         \label{Log-likelihoods (IMDB)}
     \end{subfigure}
     \hfill
     \begin{subfigure}[b]{0.49\textwidth}
         \centering
         \includegraphics[width=\textwidth]{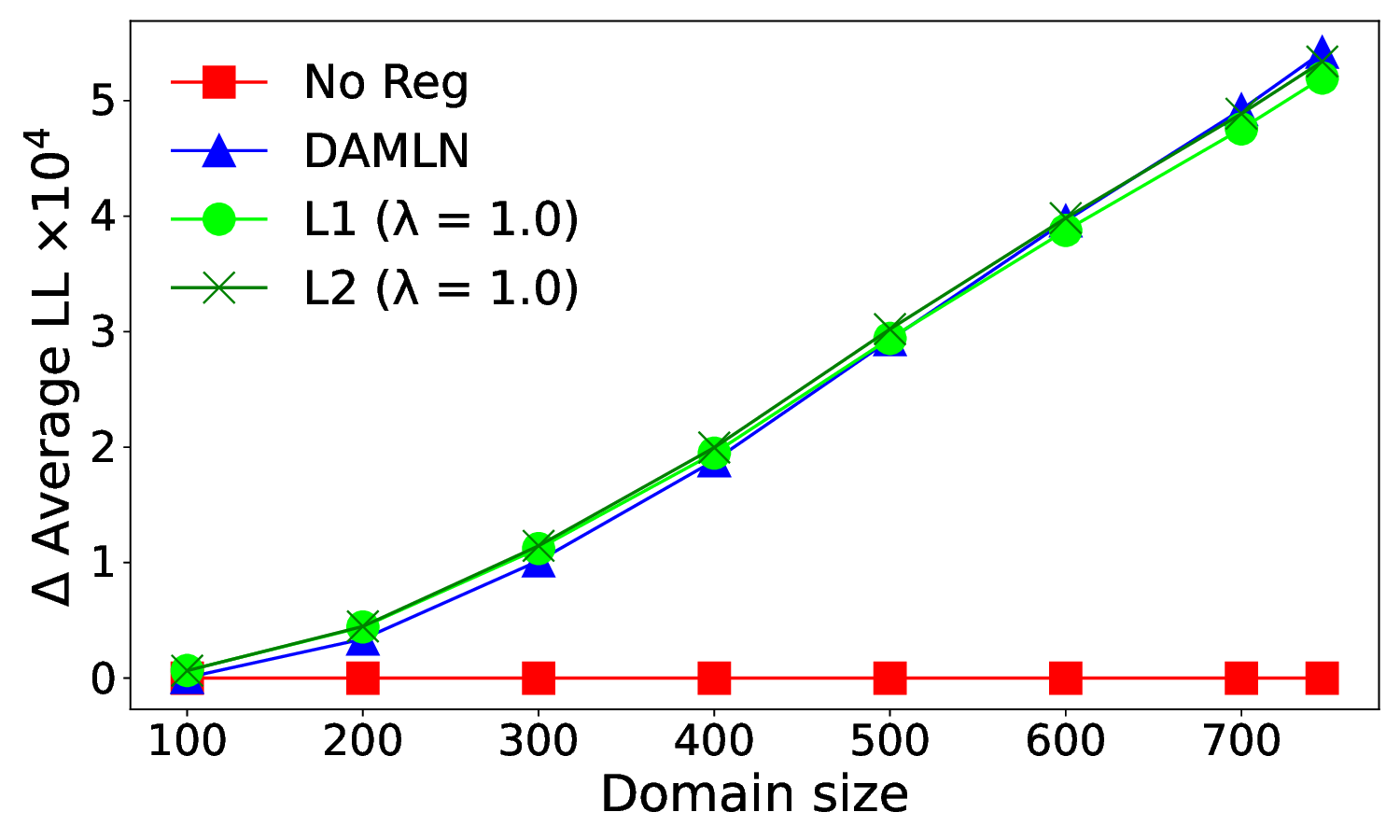}
         \caption{$\Delta$-log-likelihoods (WebKB)}
         \label{Log-likelihoods (WebKB)}
     \end{subfigure}
     \hfill
     \begin{subfigure}[b]{0.49\textwidth}
         \centering
         \includegraphics[width=\textwidth]{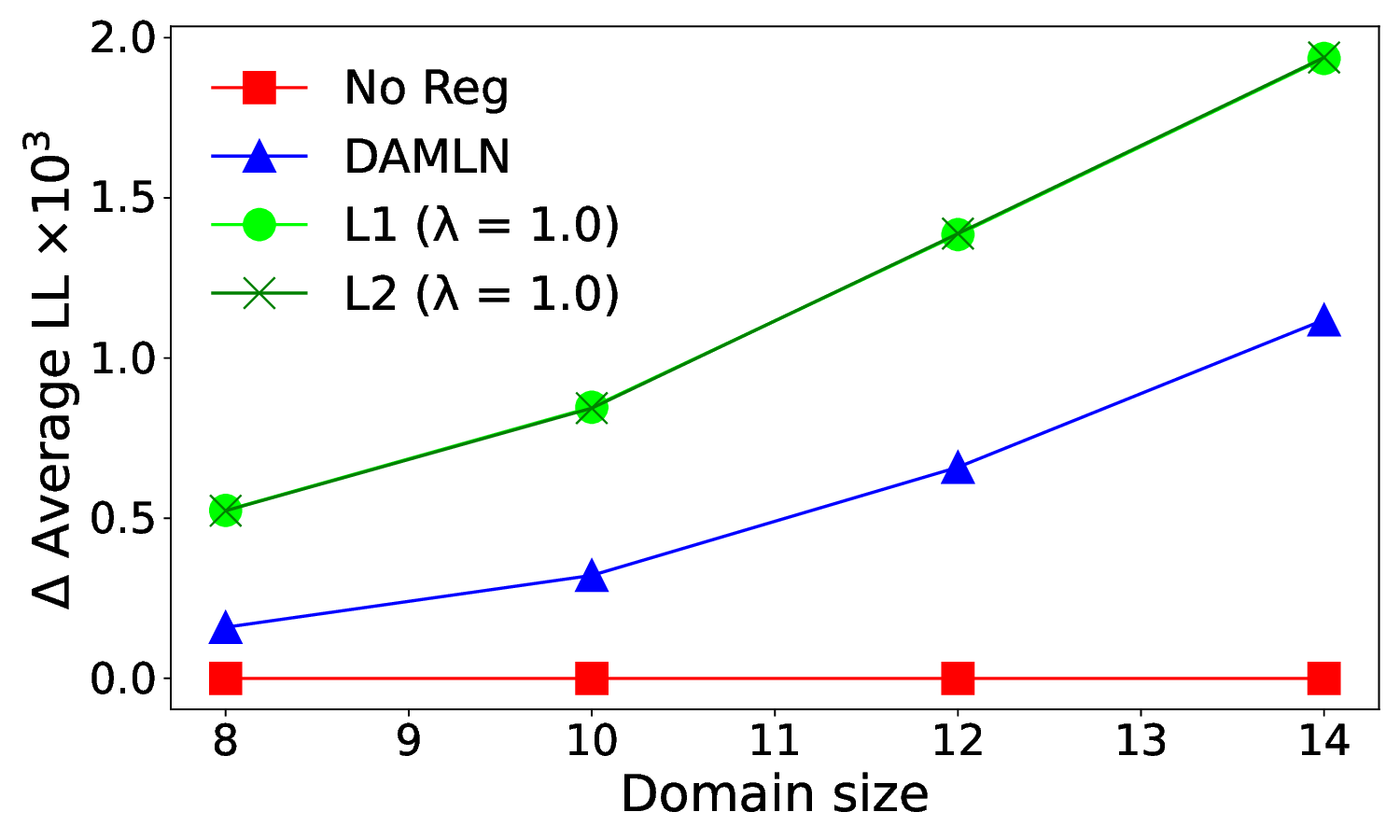}
         \caption{$\Delta$-log-likelihoods (Nations)}
         \label{Log-likelihoods (Nations)}
     \end{subfigure}
        \caption{Results for the Friends \& Smokers, IMDB, WebKB, and Nations datasets (Larger values are better)}
        \label{fig: results}
\end{figure}

\noindent 
Figure \ref{fig: results} shows the difference between the average log-likelihood obtained with the regularization approaches and the one obtained without regularization. For each of the four datasets, methods that reduce parameter variance consistently improve target set likelihood by several orders of magnitude (except for L2 regularization on FS). This effect is more pronounced as the target set size grows. 
Among the methods that reduce parameter variance, L1 and L2 regularization have similar performances. DA-MLNs outperform L1 and L2 on the FS dataset, but underperform on Nations and IMDB, while producing similar results on WebKB. 
Note that MLNs learned with L1 and L2 regularization are not domain-aware and work with the same parameters across domain sizes. Thus, it is unclear whether domain-aware parameter variance reduction methods are generally preferable to domain-unaware methods in practice. This can be observed in the relative under-performance of DA-MLNs on some of the datasets. 




\section{Conclusion}

In this paper, we analyze the generalization behavior of Markov Logic Networks when used across domain sizes. We observe that, unlike independent and identically distributed data, relational data does not admit consistency of parameter estimation. We then formalize this inconsistency in terms of the different (and mutually inconsistent) notions of maximum likelihood estimation for the weights of an MLN, when only partial data is observed. In our main theoretical result, we characterize conditions based on the parameter variance of the MLN that minimize this inconsistency. These theoretical conditions motivate and justify weight-learning approaches that decrease parameter variance. To empirically verify these claims we evaluate the generalization performance of three approaches that reduce parameter variance: L1 and L2 regularization, and Domain-Size Aware Markov Logic Networks. Our findings validate that reducing parameter variance consistently improves dataset-likelihoods over larger domains. 



\section{Acknowledgments}
SM thanks Kilian Rückschlo{\ss}  for pointing towards the problem investigated in this paper.



\bibliographystyle{splncs04}
\bibliography{main}

\section*{Appendix}

\begin{proof} [the lower bound proof for Proposition \ref{prop: Z factorisation}]
\begin{align*}
Z(n+m) &=  \sum_{\omega } w(\omega) \\
&\geq \sum_{\omega} w(\omega \downarrow [n]) \times w(\omega \downarrow [\bar{n}]) \times \prod_{d \in [k]} (w^{min}_{d})^{ \binom{n+m}{d} - \binom{n}{d} - \binom{m}{d}} \\
&= \sum_{\omega} w(\omega \downarrow [n]) \times w(\omega \downarrow [\bar{n}]) \times M_{min} \\
&= M_{min}\sum_{\substack{\omega' \in \Omega^{(n)} \\ \omega'' \in \Omega^{(m)}}} C_{n,m} \times w(\omega') \times w(\omega'')  \\
&= M_{min}C_{n,m} Z(n)Z(m) 
\end{align*}
\qed
\end{proof}

\begin{proof} [the lower bound proof for Theorem \ref{th: inequality_bounds}]
\begin{align*}
      P_{\Phi}^{(n+m)}\downarrow [n] (\omega') &= \sum_{\substack{  \omega \in \Omega^{(n+m)} \\ \omega \downarrow [n] = \omega'}} \frac{w(\omega)}{Z(n+m)}
\end{align*}
Using Proposition \ref{prop: Z factorisation}, we have:
\begin{align*}
      P_{\Phi}^{(n+m)}\downarrow [n] (\omega') &\geq \frac{1}{Z(n)Z(m)M_{max}C_{n,m}} \sum_{\substack{  \omega \in \Omega^{(n+m)} \\ \omega \downarrow [n] = \omega'}} w(\omega)
\end{align*}
Using Proposition \ref{prop: factorising_interpetation}, we have:
\begin{align*}
      P_{\Phi}^{(n+m)}\downarrow [n] (\omega') &\geq   \frac{1}{Z(m)M_{max} C_{n,m}} \sum_{\substack{  \omega \in \Omega^{(n+m)} \\ \omega \downarrow [n] = \omega'}} \frac{w(\omega') w(\omega \downarrow [\bar{n}]) M_{min}}{Z(n)}\\
      & =   \frac{1}{Z(m)M_{max}C_{n,m}} w(\omega')\sum_{\substack{  \omega \in \Omega^{(n+m)} \\ \omega \downarrow [n] = \omega'}} \frac{w(\omega \downarrow [\bar{n}]) M_{min}}{Z(n)}\\
      & =  \frac{1}{Z(m)M_{max} C_{n,m}} w(\omega') \frac{ Z(m) C_{n,m} M_{min}}{Z(n)}\\
      &=  \frac{M_{min}}{M_{max}}  P^{(n)}_{\Phi}(\omega')
  \end{align*}
\qed
\end{proof}
\newpage
\begin{proof} [of Corollary \ref{cor: log-likelihood + sigma}]\\
\\
\text{Using the bound derived in Theorem \ref{th: inequality_bounds}, we have:}
\begin{align*}
     \Delta^{-1} \times P_{\Phi}^{(n)}(\omega') &\leq P_{\Phi}^{(n+m)}\downarrow [n] (\omega')\\
     \bigl(P_{\Phi}^{(n+m)}\downarrow [n] (\omega')\bigr)^{-1} &\leq \Delta \times \bigl(P_{\Phi}^{(n)}(\omega')\bigr)^{-1}\\
     -\log P_{\Phi}^{(n+m)}\downarrow [n](\omega) &\leq -\log P_{\Phi}^{(n)}(\omega)  + \log\Delta
\end{align*}
\qed
\end{proof}

\begin{proof} [of Corollary \ref{cor: kl-divergence and likelihood}]\\
\\
\text{Using Theorem \ref{th: kl-divergence}, we have:}
\begin{align*}
KL(P_{\Phi}^{(n+m)}\downarrow [n] || P^{(n)}_{\Phi}) &\leq \log \Delta\\
-\log P^{(n)}_{\Phi}(\omega) +  KL(P_{\Phi}^{(n+m)}\downarrow [n] || P^{(n)}_{\Phi}) &\leq -\log P^{(n)}_{\Phi}(\omega)  + \log \Delta 
\end{align*}
\qed
\end{proof}
\end{document}